# UzBERT: pretraining a BERT model for Uzbek

**B. Mansurov** and **A. Mansurov**
Copper City Labs
{b,a}mansurov@coppercitylabs.com

August 22, 2021

**Abstract**

Pretrained language models based on the Transformer architecture have achieved state-of-the-art results in various natural language processing tasks such as part-of-speech tagging, named entity recognition, and question answering. However, no such monolingual model for the Uzbek language is publicly available. In this paper, we introduce UzBERT, a pretrained Uzbek language model based on the BERT architecture. Our model greatly outperforms multilingual BERT on masked language model accuracy. We make the model publicly available under the MIT open-source license.

**Keywords**: Uzbek language, Cyrillic script, pretrained language model, BERT, natural language processing

## 1 Introduction

In natural language processing (NLP), Transformer-based [Vaswani et al. 2017] pretrained language models have achieved state-of-the-art results on a wide range of tasks. Such models are publicly available for high-resource languages, e.g., BERT [Devlin et al. 2018] and RoBERTa [Liu et al. 2019] for English, and CamemBERT [Martin et al. 2019] for French. However, there is no such public monolingual model for the low-resource Uzbek language.

Transformer-based multilingual models such as multilingual BERT [Devlin et al. 2018], XLM [Lample and Conneau 2019], and XLM-R [Conneau et al. 2019] have been trained on multiple languages (including Uzbek) with the hope of transferring knowledge from resource-rich languages to low-resource languages. These multilingual models demonstrate convincing results in zero-shot cross-lingual model transfer [Pires et al. 2019], but they underperform their monolingual counterparts [Martin et al. 2019; Virtanen et al. 2019] on downstream tasks. Moreover, multilingual models contain larger vocabularies and more parameters[1] than monolingual models, thus requiring high-memory GPUs in order to fine-tune them. As a result, monolingual models in various languages [Virtanen et al. 2019; Lee et al. 2020] have been pretrained and made publicly available.

---
[1]The vocabulary size of monolingual BERT is 30K, while it is 119K for multilingual BERT. Also monolingual BERT has 110M parameters, and multilingual BERT has 168M parameters.

In this paper, we introduce the first publicly available Uzbek model based on the BERT architecture. Uzbek is a low-resource language — it lacks publicly available language models, labeled datasets, or even large amounts of raw text. In order to build the model, we first develop a high-quality news corpus consisting of ~142M words. We then pretrain a model, and call it UzBERT. We evaluate our model's performance against multilingual BERT (mBERT) on masked language model accuracy.[2] Our comparisons show that UzBERT achieves far superior results than mBERT on this metric. We make our model publicly available[3] under the MIT open-source license to encourage research and application development for the Uzbek language.

## 2 Background

**Uzbek language**

Uzbek is a morphologically rich, agglutinative language in the Turkic language family. It is the national language of Uzbekistan and has two active writing systems — Cyrillic and Latin. According to Togʻayev et al. 1999, the Cyrillic script was introduced in 1956, and the Latin script was (re-)introduced in 1995.

Although the Explanatory Dictionary of the Uzbek Language [Mirzayev et al. 2006] contains over 80K words and expressions, many more words can be formed by affixation. Even sentences in English can be expressed as single words in Uzbek. For example, "Are you one of those who won?" can be translated as "Ютганларданмисиз?". Here is how the stem and suffixes of the Uzbek word correspond to the English words: `ют` (won) `ган` (who) `лар` (those) `дан` (one of) `ми` (are) `сиз` (you).

Uzbek morphology is so rich that words with infinite number of letters can be formed. For example, the word `уйдагилар` means "those, who/that are at home", and `уйдагилардигилар` means "those in the possession of those, who/that are at home", while `уйдагилардигилардагилар` means "those in the possession of those, who/that are in the possession of those, who/that are at home". We can keep adding the suffix `дагилар` to create longer and longer words. At some point, it becomes hard to understand the meaning of the word, but nevertheless, the Uzbek morphology allows for such constructions.

On the other hand, Uzbek is a resource-poor language. Very few public datasets are available for NLP research in the language. Two of the main resources are the

---

[2]Because of the lack of public datasets for part-of-speech tagging, named entity recognition, and other NLP tasks, we haven't yet been able to evaluate UzBERT on these tasks.

[3]You may download UzBERT from `https://huggingface.co/coppercitylabs/uzbert-base-uncased`.

Uzbek Wikipedia and the Common Crawl corpus. According to the Wikipedia dump of 2021-07-19, there are 12.2M words[4] in the Uzbek Wikipedia. In addition to being small, the Uzbek Wikipedia suffers from low quality articles. Most articles were bulk imported from an Uzbek encyclopedia [Aminov et al. 2006], where full words are replaced by their abbreviations to save space in print. The filtered Common Crawl corpus[5] used by XLM-R [Conneau et al. 2019] is not big either — it contains ~91M words.[6] This corpus contains not only article content but also surrounding user-interface texts, making it noisy.

Even when one gathers large amounts of raw text, it is more than likely some of the data will be in Cyrillic, and the rest in Latin. In order to utilize all of the data, Cyrillic text needs to be converted into Latin, or vice versa. Mansurov and Mansurov 2021 describe an approach to transliterating between these two writing systems using machine learning. To train UzBERT we collect and use only Cyrillic texts.

**Transformer-based pretrained language models**

In NLP, traditionally, downstream tasks are solved by obtaining word embeddings [Mikolov et al. 2013; Peters et al. 2018] from raw text and feeding those embeddings into task-specific architectures. Recurrent Neural Networks (RNN) are mainly used in such architectures. A recent trend has been to pretrain a Transformer-based language model such as BERT and RoBERTa on large amounts of unannotated text, and fine-tune it for a downstream task with much less labeled data. This approach outperforms previous attempts on multiple tasks such as language understanding and question answering [Devlin et al. 2018].

The Transformer utilizes the attention mechanism without recurrence and is composed of an encoder and decoder stacks. BERT consists of an encoder stack only and is pretrained with a "masked language model" (MLM) and a "next sentence prediction" (NSP) tasks. In the MLM task, some input tokens are randomly masked, and the objective is to predict the original masked tokens. In the NSP task, given two sequences of text, the goal is to predict whether the second sequence follows the first in the original text. Once trained, the language model can be extended with an additional layer, and fine-tuned to solve a variety of downstream tasks such as part-of-speech tagging.

---

[4] An Uzbek Wikipedia dump is downloaded from `https://dumps.wikimedia.org/other/cirrussearch/20210719/uzwiki-20210719-cirrussearch-content.json.gz` and the "text" property of each JSON line is extracted. The number of words is counted with `wc -w`.

[5] `http://data.statmt.org/cc-100/uz.txt.xz`

[6] For comparison, BERT was trained on 3300M English words.

## 3  Previous Work

Attempts have been made to create word embeddings (but not Transformer-based pretrained language models) for Uzbek.

Al-Rfou et al. 2013 train embeddings for more than 100 languages (including Uzbek) using Wikipedia in those languages. Their vocabulary contains the most frequently occurring 100K words. Grave et al. 2018 create distributed word representations for 157 languages using the Wikipedia and Common Crawl corpora. Their Uzbek Wikipedia model[7] contains ~110K words, while the Common Crawl model[8] contains ~830K words. Kuriyozov et al. 2020 also develop word embeddings for Uzbek (among other Turkic languages) using fastText [Bojanowski et al. 2017] (and align these embeddings with embeddings for other Turkic languages). Their Uzbek training data contains 24M words crawled from websites [Baisa et al. 2012] and their model contains ~200K words[9].

All of the above-mentioned embeddings are trained for the Latin script of the Uzbek language. Mansurov and Mansurov 2020 develop word embeddings for the Cyrillic script using the word2vec [Mikolov et al. 2013], GloVe [Pennington et al. 2014], and fastText [Bojanowski et al. 2017] algorithms. The authors gather data by crawling websites in the "uz" domain. Their training data include more than 79M words.

The main limitation of embeddings produced by the above-mentioned models is that they are context independent — each word is assigned one vector regardless of the number of meanings it may have. Moreover, word2vec and GloVe are word-level models and cannot encode out-of-vocabulary words.

As far as we know, no Transformer-based monolingual Uzbek language model has been made publicly available. Therefore, two main contributions of this paper are gathering a high quality Cyrillic corpus and training a BERT-based model for Uzbek using this corpus.

## 4  UzBERT

**Model**

UzBERT has the same architecture (12 layers, 768 hidden dimensions, 12 attention heads, 110M parameters), training objectives (MLM and NSP), hyperparameters such as the dropout probability of 0.1 and a `gelu` activation, and the vocabulary size of 30K tokens as the original BERT.

---

[7] https://dl.fbaipublicfiles.com/fasttext/vectors-wiki/wiki.uz.vec

[8] https://dl.fbaipublicfiles.com/fasttext/vectors-crawl/cc.uz.300.vec.gz

[9] https://zenodo.org/record/3666697/files/uz.sg.300.vec.tar.gz?download=1

## Data

Our data consists of ~625K news articles that are obtained by crawling websites in the Uzbek language. The articles span across many domains such as agriculture, economics, history, literature, and law. However, almost all of them are written in the same journalistic genre. The total number of words in those articles is ~142M, of which ~140M words are used for training, and the remaining for validation. All words are down-cased and split into sub-word units using the WordPiece tokenizer [Wu et al. 2016], which we trained on the same corpus.

For evaluation, we gather a new dataset of news and encyclopedia articles (seen by neither UzBERT, nor mBERT) in both Cyrillic and Latin. Both the news and encyclopedia datasets consist of ~7K words each. We split the data into partly-overlapping sequences of 128 words and randomly mask one word in each sequence. In total, we have 872 such sequences. Similar to the Uzbek Wikipedia, the encyclopedia evaluation dataset is written in a terse style with lots of abbreviations, such as `ш.` for `шаҳар` (city) and `й.` for `йил` (year).

## Pretraining

We use the transformers library [Wolf et al. 2019] (version 4.8.2) and train a model on an NVIDIA RTX A6000 graphics card for about 2.5 days. To speed up the process, we first pretrain the model with batch size of 300 and sequence length of 128 tokens for 36 epochs. Then, to learn positional embeddings, we continue pretraining for 4 more epochs, but increase the sequence length to 512 tokens and reduce the batch size to 50.

## Experiments

We evaluate our model's performance by comparing its MLM accuracy to that of mBERT[10] using the evaluation dataset described in the previous section. We use the Cyrillic version of the dataset to evaluate UzBERT, and the Latin version to evaluate mBERT (as these models were trained on Cyrillic and Latin texts, respectively).

Ideally, we would have liked to evaluate UzBERT on a downstream task. Unfortunately, we could not find any public datasets in Uzbek suitable for such an evaluation. We hope to revisit this kind of extrinsic evaluation once we produce such a dataset.

---

[10]The mBERT model we used is available at `https://huggingface.co/bert-base-multilingual-uncased` and is trained on the top largest 102 language Wikipedias.

# 5 Results

The MLM accuracy results are shown in Table 1. The table presents the mean accuracy and standard deviation (in parentheses) of five separate tests. "Top 1 Match" means the masked word was the top suggested word by the model. Similarly, "Top 3 Match" means the masked word was among the top 3 words suggested by the model.

| Model | Evaluation dataset | Top 1 Match | Top 3 Match | Top 5 Match |
|---|---|---|---|---|
| UzBERT | News | **64.06** (1.08%) | **81.40%** (0.90%) | **85.32%** (0.83%) |
| mBERT | News | 14.52% (0.92%) | 18.81% (1.75%) | 20.32% (1.61%) |
| UzBERT | Encyclopedia | **32.25%** (0.84%) | **44.40%** (1.04%) | **48.37%** (0.84%) |
| mBERT | Encyclopedia | 14.33% (1.36%) | 19.24% (1.87%) | 20.64% (2.05%) |

Table 1: The MLM accuracy of UzBERT vs. mBERT on the news and encyclopedia evaluation datasets.

# 6 Discussion

Recall that UzBERT was trained on news articles, while mBERT was trained on Wikipedia. Consequently, UzBERT outperforms mBERT by a wide margin on the news evaluation set. It also has a higher accuracy than mBERT on the encyclopedia dataset, albeit with a reduced margin.

In our opinion, mainly three things contribute to the superior performance of UzBERT:

1. Uzbek language data used for training mBERT is about eleven times less than the amount of data used for training UzBERT.
2. UzBERT training data is of higher quality that that of mBERT. In order to collect training data for UzBERT, we hand-pick our sources, and create site-specific content extractors to extract only the article title an content.
3. Transfer learning to Uzbek from other languages may not have been successful for mBERT. We assume that the benefits of transfer learning will become apparent when mBERT is fine-tuned and evaluated on downstream tasks.

When we compare UzBERT against itself on the two evaluation datasets, the lower scores on the encyclopedia set can be attributed to the terse writing style in the evaluation data, to which UzBERT didn't have access during training. As mentioned above, the articles contain many abbreviations.

Somewhat surprisingly, the performance of mBERT is relatively the same regardless of the evaluation set. We expected it to perform better on the encyclopedia articles. It may be an indication that not enough Uzbek text was used for pretraining mBERT.

# 7 Conclusion

This research has focused on building a monolingual pretrained Uzbek language model based on the BERT architecture. The result is the first publicly available such model — UzBERT. Even though our model was trained on a small corpus (~140M words), its accuracy on masked language model is far better than that of multilingual BERT.

One of the advantages of UzBERT over mBERT is that its vocabulary size is smaller (thus it requires less resources for fine-tuning) and theoretically better captures the intricacies of the language because it was trained only on Uzbek texts. However, mBERT is more desirable in situations where task-specific fine-tuning data is available in a non-Uzbek language. In such cases mBERT can be fine-tuned in a language other than Uzbek, and tested on Uzbek texts. For that to happen, though, mBERT should be trained on a much higher quality Uzbek text than the Uzbek Wikipedia as the model's performance on MLM accuracy lags far behind UzBERT.

Future research on UzBERT should consider training a model with additional texts written in different genres. Martin et al. 2019 show that a model trained on 4GB of text is as good as a model trained on 130+GB of text. UzBERT is trained on 1.9GB of text. Doubling its size will hopefully result in a language model that is as good as a model trained on tens of GB of text.

Because of the lack of public datasets for downstream tasks in Uzbek, we were unable to test its performance on such tasks. Another direction for future work is to produce such datasets and evaluate UzBERT on downstream tasks.

Finally, it would be interesting to see the impact of the tokenizer on model's performance, similar to the work done in Lee et al. 2020 for Korean. Being a highly inflectional language, Uzbek may benefit from a tokenizer that is able to correctly split words into stems and suffixes. For example, the UzBERT tokenizer considers the word `менинг` (my) to consist of one token. Morphologically speaking, it's a word consisting of the stem `мен` (I) and the suffix `нинг` ('s). Such trivial cases can probably be solved by an unsupervised tokenizer such as WordPiece, but to correctly tokenize texts such as `Ютганларданмисиз?` (Are you one of those who won?), we will most certainly need a carefully crafted tokenizer based on Finite State Machines, for example.

We hope that UzBERT generates more interest from scholars to the Uzbek language and serves as a catalyst for developing new resources in building robust Uzbek language models.

# 8 Acknowledgments

We would like to thank N. Mansurov for manually gathering the evaluation datasets.